%% file: neurips_2022.tex
\documentclass{article}

\newcommand{\tc}[1]{\multicolumn{1}{c}{#1}} 
\newcommand{\tr}[1]{\multicolumn{1}{r}{#1}} 
\newcommand{\summary}[1]{\vspace{0.3em} \noindent \fbox{\parbox{1\linewidth}{\emph{\textbf{Summary}: #1}}}}

\usepackage{fancyhdr}
\usepackage{caption}
\usepackage{subfigure}
\usepackage{amsmath}
\usepackage{array}
\usepackage{url}
\usepackage{tcolorbox}
\usepackage{algorithm}
\usepackage[noend]{algpseudocode}
\usepackage{calc}
\usepackage[export]{adjustbox}

\usepackage[preprint,nonatbib]{neurips_2022}

\usepackage[utf8]{inputenc} 
\usepackage[T1]{fontenc}    
\usepackage{hyperref}       
\usepackage{url}            
\usepackage{booktabs}       
\usepackage{amsfonts, amssymb}       
\usepackage{nicefrac}       
\usepackage{microtype}      
\usepackage{xcolor}         

\title{DeFL: Decentralized Weight Aggregation for Cross-silo Federated Learning}

\author{Jialiang Han\\
Key Lab of High-Confidence Software Technology, MoE (Peking University), Beijing\\
\texttt{hanjialiang@pku.edu.cn}\\
\And
Yudong Han\\
Key Lab of High-Confidence Software Technology, MoE (Peking University), Beijing\\
\texttt{hanyd@pku.edu.cn}\\
\And
Gang Huang\\
Key Lab of High-Confidence Software Technology, MoE (Peking University), Beijing\\
\texttt{hg@pku.edu.cn}\\
\And
Yun Ma\\
Institute for Artificial Intelligence, Peking University, Beijing\\
\texttt{mayun@pku.edu.cn}\\
}

\begin{document}

\maketitle

\begin{abstract}
   Federated learning (FL) is an emerging promising paradigm of privacy-preserving machine learning (ML). An important type of FL is cross-silo FL, which enables a small scale of organizations to cooperatively train a shared model by keeping confidential data locally and aggregating weights on a central parameter server. However, the central server may be vulnerable to malicious attacks or software failures in practice. To address this issue, in this paper, we propose \texttt{DeFL}, a novel decentralized weight aggregation framework for cross-silo FL. \texttt{DeFL} eliminates the central server by aggregating weights on each participating node and weights of only the current training round are maintained and synchronized among all nodes. We use \texttt{Multi-Krum} to enable aggregating correct weights from honest nodes and use \texttt{HotStuff} to ensure the consistency of the training round number and weights among all nodes. Besides, we theoretically analyze the Byzantine fault tolerance, convergence, and complexity of \texttt{DeFL}. We conduct extensive experiments over two widely-adopted public datasets, i.e. CIFAR-10 and Sentiment140, to evaluate the performance of \texttt{DeFL}. Results show that \texttt{DeFL} defends against common threat models with minimal accuracy loss, and achieves up to 100x reduction in storage overhead and up to 12x reduction in network overhead, compared to state-of-the-art decentralized FL approaches.
 
\end{abstract}

\section{Introduction} \label{Section:Introduction}
\input{Sections/Introduction.tex}

\section{Related Work} \label{Section:Related_Works}
\input{Sections/Related_Work.tex}

\section{Methodology} \label{Section:Methodology}
\input{Sections/Methodology.tex}

\section{Analysis} \label{Section:Analysis}
\input{Sections/Analysis}

\section{Evaluation} \label{Section:Evaluation}
\input{Sections/Evaluation.tex}

\section{Threats to Validity} \label{Section:Discussion}
\input{Sections/Discussion.tex}

\section{Conclusion} \label{Section:Conclusion}
\input{Sections/Conclusion.tex}

\bibliographystyle{plain}
\bibliography{ref}

\section*{Checklist}

\begin{enumerate}

\item For all authors...
\begin{enumerate}
  \item Do the main claims made in the abstract and introduction accurately reflect the paper's contributions and scope?
    \answerYes{}
  \item Did you describe the limitations of your work?
    \answerYes{}
  \item Did you discuss any potential negative societal impacts of your work?
    \answerNo{The data privacy and security has raised considerable concerns of end-users and this work benefits the data privacy and security of federated learning (FL).}
  \item Have you read the ethics review guidelines and ensured that your paper conforms to them?
    \answerYes{}
\end{enumerate}

\item If you are including theoretical results...
\begin{enumerate}
  \item Did you state the full set of assumptions of all theoretical results?
    \answerYes{}
        \item Did you include complete proofs of all theoretical results?
    \answerYes{}
\end{enumerate}

\item If you ran experiments...
\begin{enumerate}
  \item Did you include the code, data, and instructions needed to reproduce the main experimental results (either in the supplemental material or as a URL)?
    \answerYes{We will make the source code of this paper publicly available online when this paper is accepted.}
  \item Did you specify all the training details (e.g., data splits, hyperparameters, how they were chosen)?
    \answerYes{}
        \item Did you report error bars (e.g., with respect to the random seed after running experiments multiple times)?
    \answerYes{}
        \item Did you include the total amount of compute and the type of resources used (e.g., type of GPUs, internal cluster, or cloud provider)?
    \answerYes{}
\end{enumerate}

\item If you are using existing assets (e.g., code, data, models) or curating/releasing new assets...
\begin{enumerate}
  \item If your work uses existing assets, did you cite the creators?
    \answerYes{}
  \item Did you mention the license of the assets?
    \answerYes{}
  \item Did you include any new assets either in the supplemental material or as a URL?
    \answerYes{}
  \item Did you discuss whether and how consent was obtained from people whose data you're using/curating?
    \answerYes{}
  \item Did you discuss whether the data you are using/curating contains personally identifiable information or offensive content?
    \answerYes{}
\end{enumerate}

\item If you used crowdsourcing or conducted research with human subjects...
\begin{enumerate}
  \item Did you include the full text of instructions given to participants and screenshots, if applicable?
    \answerNA{}
  \item Did you describe any potential participant risks, with links to Institutional Review Board (IRB) approvals, if applicable?
    \answerNA{}
  \item Did you include the estimated hourly wage paid to participants and the total amount spent on participant compensation?
    \answerNA{}
\end{enumerate}

\end{enumerate}

\clearpage
\appendix
\input{Sections/Appendix.tex}

\end{document}

%% file: Sections/Introduction.tex
In many real-world scenarios, such as e-commerce, medical diagnosis, and Internet of Things (IoT), data are distributed among devices or organizations and the volume of local data is insufficient to train reliable models without over-fitting. To address this limitation, it is a common practice to feed local data into a centralized server and train a global model. Undoubtedly, it raises concerns about data ownership, privacy, security, and monopolies. Recently, emerging federated learning (FL)~\cite{DBLP:conf/aistats/McMahanMRHA17} mitigates some of these concerns by training a global model without gathering confidential data from each participating node~\cite{DBLP:conf/ccs/ShokriS15}. Cross-silo FL~\cite{DBLP:conf/usenix/ZhangLX00020, DBLP:conf/nips/MarfoqXNV20} is an important type of FL where 2-100 organizations collectively aggregate weights on a central parameter server, which is assumed to be trusted and reliable among organizations. However, this assumption may not hold in practice. For example, the central server could be malicious, leading to poisoning the model~\cite{DBLP:conf/raid/FungYB20, DBLP:conf/aistats/BagdasaryanVHES20, DBLP:conf/iclr/XieHCL20, DBLP:conf/icml/XieKG19}, or skewing the model by favoring particular clients~\cite{DBLP:series/lncs/LyuXWY20, DBLP:conf/aies/0001LLCCWNY20}. Besides, fatal crashes in central servers could lead to an accuracy drop, convergence time increase, or even training procedure abortion.

To address the preceding problems raised by the central parameter server of FL, some decentralized FL solutions are proposed by eliminating the central server. These solutions can be categorized into two directions. One direction is to dynamically elect a leader, where weights are aggregated and transmitted to other nodes~\cite{warnat2021swarm, warnat2020swarm, DBLP:journals/network/LiCLHZY21, DBLP:journals/corr/abs-2101-06905}. The leader takes the place of the central server of FL. However, if the dynamically elected leader is detected (probably through overmuch network bandwidth~\cite{DBLP:journals/corr/abs-2201-05286}) and then attacked, or if the leader behaves maliciously, the risks of model poisoning and skewing still exist. The other direction is to leverage a blockchain to maintain weights and coordinate the weight aggregation~\cite{DBLP:journals/tpds/ShayanFYB21, feng2021blockchain, DBLP:journals/network/LiCLHZY21, DBLP:conf/blockchain2/RamananN20, DBLP:journals/corr/abs-2101-06905, DBLP:conf/bigcom/BaoSXHH19, DBLP:conf/wcnc/Pokhrel020, DBLP:journals/network/LuHZMZ21, DBLP:journals/access/ToyodaZZM20, DBLP:conf/blockchain2/HarrisW19}. However, most of them are implemented based on a third-party blockchain platform, such as Ethereum or FISCO, therefore suffering from unnecessary storage and network overhead. The reason is that they maintain the consistency of all history weights due to underlying consensus mechanisms. However, FL requires weights of only the current training round for updating and does not require updates in one round to be recorded in a particular sequence. 

\begin{figure}[htbp]
  \centering
  \includegraphics[width=0.8\textwidth]{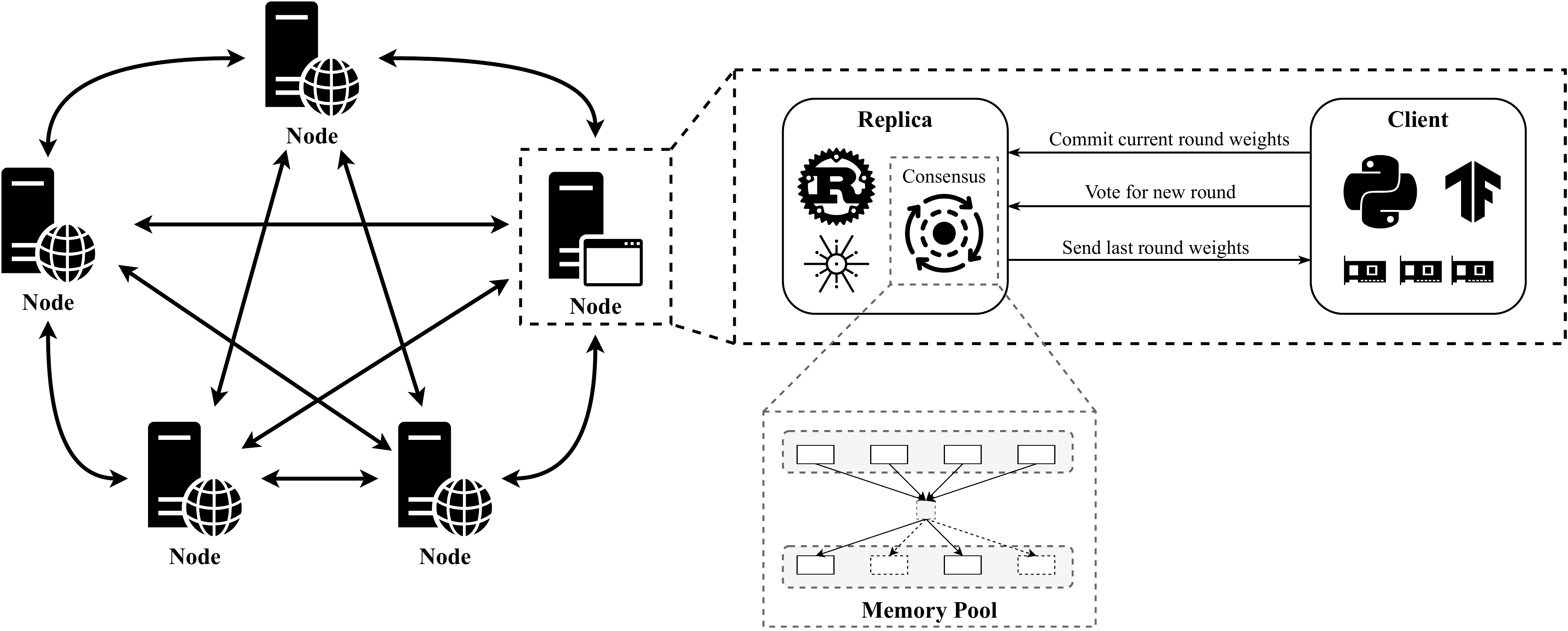}
  \caption{The architecture of \texttt{DeFL}}
  \label{architecture}
\end{figure}

To this end, in this paper, we propose \texttt{DeFL}, a novel decentralized weight aggregation framework for cross-silo FL. The key idea of \texttt{DeFL} is in two folds. First, the local updates of all nodes are aggregated on each node so that the reliability concern of a leader or central server can be mitigated. Second, weights of only the current training round are maintained and synchronized so that the storage and network overhead can be significantly reduced. Realizing such an idea faces two challenges. First, the weights aggregated on or updated (local trained) by faulty or adversarial nodes are not reliable. Second, synchronizing weights of only the current round brings in inconsistency in the round number $\mathit{round\_id}$. For example, at some specific point in the training procedure, $\mathit{round\_id}_A$ on client $A$ with sufficient computation and network resources could be larger than $\mathit{round\_id}_B$ on client $B$ with insufficient resources. As a result, aggregating weights of $\mathit{round\_id}_A$ and $\mathit{round\_id}_B$ is inconsistent with \texttt{FedAvg}~\cite{DBLP:conf/aistats/McMahanMRHA17} in the standard FL setting, leading to accuracy drop or convergence time increase. To tackle these challenges, we abstract each participating node as two roles, i.e. a client and a replica, as shown in Figure~\ref{architecture}. A client enables aggregating \textit{correct} weights from honest nodes with a weight filter based on \texttt{Multi-Krum}~\cite{DBLP:conf/nips/BlanchardMGS17}. A replica ensures the consistency of $\mathit{round\_id}$ and weights of the current and last rounds with a synchronizer based on \texttt{HotStuff}~\cite{DBLP:conf/podc/YinMRGA19}. 

To evaluate the performance of \texttt{DeFL}, we conduct extensive experiments over two widely-adopted public datasets, i.e. CIFAR-10 and Sentiment140. We comprehensively measure Byzantine fault tolerance and scalability on the accuracy, computational overhead, storage overhead, and network overhead. The results show that \texttt{DeFL} defends against common threat models with minimal accuracy loss, and achieves up to 100x reduction in storage overhead and up to 12x reduction in network overhead, compared to state-of-the-art decentralized FL baselines. We will make the source code of this paper publicly available online when this paper is accepted. Our contributions of this paper can be summarized as follows.
\begin{itemize}
    \item We propose a novel decentralized weight aggregation framework for cross-silo FL (\texttt{DeFL}), where weights are aggregated on each node and weights of only the current training round are maintained and synchronized.
    \item We design a weight filter based on \texttt{Multi-Krum} to enable aggregating correct weights and design a synchronizer based on \texttt{HotStuff} to ensure consistency of $\mathit{round\_id}$ and weights.
    \item We theoretically analyze the Byzantine fault tolerance, convergence, and overhead of \texttt{DeFL}.
    \item We use public datasets to measure the performance of \texttt{DeFL} and demonstrate its superior to state-of-the-art baselines. 
\end{itemize}


%% file: Sections/Related_Work.tex
In this section, we introduce related work about federated learning (FL) and decentralized FL.

\textbf{Federated Learning.} To allow users to collectively reap the benefits of shared models trained from rich data without transmitting raw data, McMahan \textit{et al.} propose FL~\cite{DBLP:conf/aistats/McMahanMRHA17}. To improve communication efficiency, Kone{\v{c}}n{\`y} \textit{et al.}~\cite{DBLP:journals/corr/KonecnyMRR16} propose structured update and sketched update. Bonawitz \textit{et al.}~\cite{DBLP:conf/mlsys/BonawitzEGHIIKK19} introduce the protocol of FL, detailed system design on devices and servers, and some specific challenges of implementation. Yang \textit{et al.}~\cite{DBLP:journals/tist/YangLCT19} categorize FL into Horizontal FL, Vertical FL, and Federated Transfer Learning. In FL settings, the central parameter server aggregates updates from devices with weights proportional to the size of local datasets, i.e. \texttt{FedAvg}~\cite{DBLP:conf/aistats/McMahanMRHA17}. Therefore, the stability, fairness, and security of the central server are crucial to FL.

\textbf{Decentralized Federated Learning.} To mitigate security and fault tolerance concerns of FL, decentralized FL~\cite{lalitha2018fully, DBLP:journals/corr/abs-1901-11173} is proposed, where users update their beliefs by aggregating information from their one-hop neighbors. The following literature focuses on aggregating weights of all active nodes and is categorized into two directions.

One direction is to dynamically elect a leader, where weights are aggregated and transmitted to other clients. For example, Swarm Learning (SL)~\cite{warnat2021swarm, warnat2020swarm} combines decentralized hardware infrastructures, and distributed machine learning (ML) with a permissioned blockchain to securely onboard members, dynamically elect the leader, and merge model parameters. BLADE-FL~\cite{DBLP:journals/corr/abs-2101-06905} allows each client to broadcast the trained model to other clients, aggregate its model with received ones, and then compete to generate a block before local training of the next round. In this direction, the network bandwidth of leaders tends to be significantly higher than other nodes~\cite{DBLP:journals/corr/abs-2201-05286} and easy to detect. If the leader is attacked or behaves maliciously, these approaches become invalid.

The other direction is to leverage a blockchain to maintain weights and coordinate the model aggregation. For example, Biscotti~\cite{DBLP:journals/tpds/ShayanFYB21} uses blockchain and cryptographic primitives to coordinate a privacy-preserving ML process. BAFFLE~\cite{DBLP:conf/blockchain2/RamananN20} uses Smart Contracts (SC) to coordinate the round delineation, model aggregation, and update tasks in FL. BFLC~\cite{DBLP:journals/network/LiCLHZY21} uses blockchain for global model storage and local model update exchange and devises an innovative committee consensus mechanism to reduce consensus computing and malicious attacks. BFL~\cite{DBLP:conf/wcnc/Pokhrel020} designs a novel reward and develops a mathematical framework that features the controllable network and parameters for privacy-aware and efficient on-vehicle FL. Lu \textit{et al.}~\cite{DBLP:journals/network/LuHZMZ21} integrate blockchain for maintaining parameters and use reinforcement learning (RL) to solve the resource sharing task. However, most of them are implemented based on a third-party blockchain platform, therefore suffering from unnecessary storage and network overhead.

%% file: Sections/Methodology.tex
\algblock{On}{EndOn}
\algnewcommand\algorithmicas{\textbf{on}}
\algrenewtext{On}[1]{\algorithmicas\ #1}
\algtext*{EndOn}

In this section, we describe the threat models, weight filtering and aggregation on a client, $\mathit{round\_id}$ and weight synchronization among replicas, and the decoupling-storage-and-consensus design of \texttt{DeFL}.

\subsection{Threat Model} \label{Section:Methodology:Threat_model}
Following existing Byzantine fault tolerant (BFT) FL approaches~\cite{DBLP:journals/tpds/ShayanFYB21, DBLP:journals/tvt/ChenCZW21, DBLP:journals/sensors/WangT22, DBLP:conf/globecom/OtoumRM20}, we assume the number of participating nodes is $n>3f$, where $f$ denotes the number of Byzantine nodes while other nodes are honest. In cross-silo FL, where device heterogeneity is relatively low, we assume partially synchronous communication~\cite{DBLP:journals/jacm/DworkLS88}, whereby all honest nodes complete a local training round before a \textit{global stabilization time of local training} (GST\_LT). $f$ Byzantine nodes are faulty or adversarial. Faulty nodes may not always commit transactions promptly. Adversarial nodes can perform three representative poisoning weight attacks, i.e., Gaussian attack~\cite{DBLP:conf/uss/FangCJG20}, sign-flipping attack~\cite{DBLP:conf/aaai/LiXCGL19} and label-flipping attack~\cite{DBLP:conf/icml/BiggioNL12}. Adversarial nodes can also commit update transactions with weights of the wrong round or commit aggregation transactions before GST\_LT. Because one node plays the role of both a client and a replica, malicious behaviors are not assumed to exist between the client and replica of the same node.

\subsection{Weight Filter}
A client enables aggregating correct weights with a weight filter based on \texttt{Multi-Krum}~\cite{DBLP:conf/nips/BlanchardMGS17}. The main idea of \texttt{Krum} is that when data are independent and identically distributed (i.i.d.), the gradient updates from honest clients tend to be close to the correct gradient, while those from the Byzantine clients tend to be arbitrary and are supposed to be omitted for aggregation. Specifically, \texttt{Krum} selects the gradient that minimizes the sum of squared distances to its $n-f$ closest gradients. \texttt{Multi-Krum} interpolates between \texttt{Krum} and \texttt{FedAvg}, thereby allowing to mix the BFT properties of \texttt{Krum} with the convergence speed of \texttt{FedAvg}. Specifically, \texttt{Multi-Krum} applies \texttt{FedAvg} on top $k$ gradients that minimizes the sum of squared distances to its $n-f$ closest gradients.

\begin{algorithm}
    \caption{Local training and weight aggregation on a client}
    \label{alg1}
    \begin{algorithmic}[1]
        \If{$\mathit{l\_round\_id}\leq \mathit{r\_round\_id}$}
            \State $\mathit{start\_time}\gets \mathit{clock}()$
            \State $\mathit{weight}_\mathit{agg}\gets \texttt{Multi-Krum}(W_1^\mathit{LAST},\dots,W_n^\mathit{LAST})$
            \State $\mathit{weight}_\mathit{new}\gets \texttt{local\_train}(\mathit{weight}_\mathit{agg}, \mathit{l\_data})$
            \State $\mathit{resp}\gets$ \texttt{commit} $\mathbf{TX}(\texttt{"UPD"}, \mathit{id}, \mathit{r\_round\_id}+1,\mathit{weight}_\mathit{new})$
            \If{$\mathit{resp}=\texttt{OK}$}
                \State $\mathit{l\_round\_id}\gets \mathit{r\_round\_id}+1$
            \EndIf
            \State $\mathit{end\_time}\gets \mathit{clock}()$
            \State $\texttt{sleep}(\max{(0, \mathrm{GST\_LT} - (\mathit{end\_time} - \mathit{start\_time}))})$
            \State \texttt{commit} $\mathbf{TX}(\texttt{"AGG"}, \mathit{l\_round\_id})$
        \EndIf
    \end{algorithmic}
\end{algorithm}
Algorithm~\ref{alg1} describes how a client executes writing operations to local data structures that do not require synchronization, i.e. local round number and local weights. When the local training round $\mathit{l\_round\_id}$ falls behind the global (replica) training round $\mathit{r\_round\_id}$, the client is supposed to update local $\mathit{round\_id}$ and local weights through local training. Within GST\_LT, the client filters and aggregates weights of other clients of the last round in Line 3, trains aggregated weights with local data in Line 4, and commits an \texttt{UPD} transaction (TX) to replicas in Line 5. Then, the client waits until GST\_LT and commits an \texttt{AGG} transaction to replicas in Line 9-10.

\subsection{Synchronizer}
A replica ensures the consistency of $\mathit{round\_id}$ and weights of the current and last round with a synchronizer based on \texttt{HotStuff}~\cite{DBLP:conf/podc/YinMRGA19}. \texttt{HotStuff}~\cite{DBLP:conf/podc/YinMRGA19} is proposed to address scaling challenge in Practical Byzantine Fault Tolerance (PBFT)~\cite{DBLP:conf/osdi/CastroL99}. Under the partially synchronous communication model~\cite{DBLP:journals/jacm/DworkLS88}, whereby a known bound $\Delta$ on message transmission holds after some unknown \textit{global stabilization time} (GST), and $n\geq 3f+1$, \texttt{HotStuff} is a leader-based BFT State Machine Replication (SMR) protocol which achieves \textit{linear} view change and optimistic responsiveness by adding a \texttt{PRE-COMMIT} phase to each view in PBFT. Its overall communication complexity per view is $O(n)$, which enables large-scale deployment on FL devices.

\begin{algorithm}
    \caption{Synchronization among replicas}
    \label{alg2}
    \begin{algorithmic}[1]
        \On{execution of $\mathbf{TX}(\texttt{"UPD"}, \mathit{id}, \mathit{target\_round\_id}, \mathit{weight})$}
            \If{$\mathit{target\_round\_id}=\mathit{r\_round\_id}+1$}
                \State $W_\mathit{id}^\mathit{CUR}\gets \mathit{weight}$
                \State \textbf{respond} \texttt{OK}
            \Else
                \State \textbf{respond} \texttt{AlreadyUPDError}
            \EndIf
        \EndOn
        \On{execution of $\mathbf{TX}(\texttt{"AGG"}, \mathit{target\_round\_id})$}
            \If{$\mathit{target\_round\_id}=\mathit{r\_round\_id}+1$}
                \State $\mathit{votes}\gets \mathit{votes}+1$
                \If{$\mathit{votes}$ meets a quorum}
                    \State $\mathit{r\_round\_id}\gets \mathit{target\_round\_id}$
                    \State $\mathit{votes}\gets 0$
                    \For{$i\gets 1$ to $n$}
                        \State $W_i^\mathit{LAST}\gets W_i^\mathit{CUR}$
                        \State $W_i^\mathit{CUR}\gets \varnothing$
                    \EndFor
                    \State \textbf{respond} \texttt{OK}
                \Else 
                    \State \textbf{respond} \texttt{NotMeetQuorumWarning}
                \EndIf
            \Else 
                \State \textbf{respond} \texttt{AlreadyAGGError}
            \EndIf
        \EndOn
    \end{algorithmic}
\end{algorithm}
Algorithm~\ref{alg2} describes how a replica executes writing operations to global data structures that require synchronization, i.e. $\mathit{round\_id}$ and weights of the current and last rounds. When executing transactions, the replica is supposed to verify that the committed $\mathit{round\_id}$ is consistent with the current global training round, in Line 2 and Line 8. When executing the \texttt{UPD} transaction, the replica synchronizes the weights of the current round of the corresponding client in Line 3. When executing the \texttt{AGG} transaction, the replica waits until the number of received \texttt{AGG} transactions meets a $quorum$, i.e. $f+1$, in Line 8-10. 
Then, the replica updates and synchronizes $\mathit{round\_id}$ and the weights of the current and last rounds in Line 11-16. 

\subsection{Decoupling Storage and Consensus}
We construct \texttt{DeFL} in two layers: a communication layer and a storage layer. In the communication layer, processes reliably broadcast their proposals and reach consensus, following the schedule of \texttt{HotStuff}. In the storage layer, weights are stored in a trusted memory pool and can be retrieved by a unique index, which does not require any extra communication.

%% file: Sections/Analysis.tex
In this section, we analyze Byzantine fault tolerance, convergence, and overhead of \texttt{DeFL}.

\subsection{Byzantine Fault Tolerance}\label{Section:Analysis:BFT}
\textbf{Lemma 1.} \textit{In} \texttt{HotStuff}\textit{, when $n\ge 3f+1$, if $t_1$ and $t_2$ are conflicting transactions, then they cannot be both committed, each by an honest replica.}

This lemma is proven in \texttt{HotStuff}~\cite{DBLP:conf/podc/YinMRGA19}. It means that all honest nodes receive transactions with the same content and in the same sequence. Therefore, each honest node receives the same weights in each round and behaves identically the same when aggregating weights. Therefore, we can treat each honest node as a parameter server and consider one honest node to represent all.

\textbf{Lemma 2.} \textit{In} \texttt{Krum}\textit{, let $V_1,...,V_n$ be any independent and identically distributed (i.i.d.) random $d$-dimensional vectors s.t. $V_i\sim G$, with $\mathbb{E}G=g$ and $\mathbb{E}\|G-g\|^2=d\sigma^2$. Let $B_1,...,B_f$ be any $f$ random vectors, possibly dependent on the $V_i$'s. If $n>2f+2$ and $\eta(n,f)\sqrt{d}\cdot\sigma<\|g\|$, where}
\begin{equation}
    \eta(n,f)\stackrel{\mathrm{def}}{=}\sqrt{2\left(n-f+\frac{f\cdot(n-f-2)+f^{2}\cdot(n-f-1)}{n-2 f-2}\right)}=\begin{cases}
    O(n), & f=O(n)\\
    O(\sqrt{n}), & f=O(1)
\end{cases},
\label{Equation:eta}
\end{equation}
\textit{then the} \texttt{Krum} \textit{function} \texttt{KR} \textit{is $(\alpha, f)$-Byzantine fault tolerant where $0\leq\alpha<\pi/2$ is defined by}
\begin{equation}
    \sin\alpha=\frac{\eta(n, f)\cdot\sqrt{d}\cdot\sigma}{\|g\|}.
\end{equation}

The definition of $(\alpha, f)$-Byzantine fault tolerance and the proof of this lemma is in \texttt{Multi-Krum}~\cite{DBLP:conf/nips/BlanchardMGS17}.

\textbf{Theorem 1.} \textit{In} \texttt{DeFL}\textit{, when $n\ge 3f+3$ and $\eta(n,f)\sqrt{d}\cdot\sigma<\|g\|$, there exists $\alpha$ such that} \texttt{DeFL} \textit{is $(\alpha, f)$-Byzantine fault tolerant.}

\textit{Proof.} Let $f_H\leq f$ be faulty nodes that fail to update weights before GST\_LT, $f_K\leq f$ be adversarial nodes that update poisoned weights before GST\_LT, where $f_H+f_K=f$. Obviously, active nodes $n_K$ participating in weight synchronization satisfy $2f+3\leq n_K\leq n$. According to Equation~\ref{Equation:eta}, $\eta(n,f)$ monotonically increases with $n$, therefore, $\eta(n_K,f_K)\sqrt{d}\cdot\sigma\leq \eta(n,f)\sqrt{d}\cdot\sigma\leq\|g\|$. Therefore, there exists $\alpha_0$ such that
\begin{equation}
    \sin\alpha_0=\frac{\eta(n_K, f_K)\cdot\sqrt{d}\cdot\sigma}{\|g\|}\leq 1,
\end{equation}
and the \texttt{Krum} function \texttt{KR} is $(\alpha_0, f_K)$-BFT. Let $f_K=f$, then \texttt{DeFL} is $(\alpha_0, f)$-BFT.

\subsection{Convergence}\label{Section:Analysis:Convergence}
Following \texttt{HotStuff}~\cite{DBLP:conf/podc/YinMRGA19}, we analyze the convergence of the Stochastic Gradient Descent (SGD) using \texttt{DeFL}. The SGD optimization is formulated as $w_{t+1}=w_t-\gamma_t\cdot\texttt{KR}(V_1^t,...,V_n^t),$. For an honest node $i,V_i^t=G(w_t,\xi_i^t)$ where $G$ is the gradient estimator and $\xi_i^t$ is a mini-batch of samples. The local standard deviation $\sigma(w)$ is defined as $d\cdot \sigma^2(w)=\mathbb{E}\|G(w,\xi)-\nabla Q(w)\|^2,$ where $Q(w)$ is the loss function.

\textbf{Lemma 3.} \textit{In} \texttt{HotStuff}\textit{, after GST, there exists a bounded time period $T_f$ such that if all honest nodes remain in view $v$ during $T_f$ and the leader for view $v$ is honest, then a decision is reached.}

This lemma is proven in \texttt{HotStuff}~\cite{DBLP:conf/podc/YinMRGA19}. It means that synchronization among replicas would only take a bounded time period and have bounded influence on convergence. 

\textbf{Lemma 4.} \textit{In} \texttt{Krum}\textit{, we assume that (i) the loss function $Q$ is three times differentiable with continuous derivatives, and is non-negative; (ii) the learning rates satisfy $\sum_{t}\gamma_{t}=\infty$ and $\sum_{t}\gamma_{t}^{2}<\infty$; (iii) the gradient estimator satisfies $\mathbb{E}G(w,\xi)=\nabla Q(w)$ and $\forall r\in\{2,...,4\},\mathbb{E}\|G(w,\xi)\|^r\leq A_r+B_r\|w\|^r$ for some constants $A_r, B_r$; (iv) there exists a constant $0\leq\alpha<\pi/2$ such that for all $w,\eta(n,f)\cdot\sqrt{d}\cdot\sigma(w)\leq\|\nabla Q(w)\|\cdot\sin\alpha$; (v) beyond a certain horizon, $\|w\|^2\geq D$, there exist $\epsilon>0$ and $0\leq\beta<\pi/2-\alpha$ such that $\|\nabla Q(w)\|\geq\epsilon>0$ and $\frac{\langle w,\nabla Q(w)\rangle}{\|w\|\cdot\|\nabla Q(w)\|}\geq\cos\beta.$ Then the sequence of gradients $\nabla Q(w_t)$ converges almost surely to zero.}

This lemma is proven in \texttt{Multi-Krum}~\cite{DBLP:conf/nips/BlanchardMGS17}. 

\textbf{Theorem 2.} \textit{In} \texttt{DeFL}\textit{, when $n\ge 3f+3$, under the five assumptions of Lemma 4, the sequence of gradients $\nabla Q(w_t)$ converges almost surely to zero.}

\textit{Proof.} Note that \texttt{DeFL} influences only \textit{Assumption (iv)} in \textit{Lemma 4} by changing the constraint that $n\ge 3f+3, f_H+f_K=f,$ and $2f+3\leq n_K\leq n$. Due to \textit{Assumption (iv)} in \textit{Theorem 2}, there exists a constant $0\leq\alpha<\pi/2$ such that for all $w,\eta(n,f)\cdot\sqrt{d}\cdot\sigma(w)\leq\|\nabla Q(w)\|\cdot\sin\alpha$. Due to the monotonicity of $\eta(n,f)$, $\eta(n_K,f_K)\cdot\sqrt{d}\cdot\sigma(w)\leq\eta(n,f)\cdot\sqrt{d}\cdot\sigma(w)\leq\|\nabla Q(w)\|\cdot\sin\alpha$, which satisfies \textit{Assumption (iv)} in \textit{Lemma 4}. According to \textit{Lemma 4}, $\nabla Q(w_t)$ converges almost surely to zero in \texttt{DeFL}.

The Byzantine fault tolerance and convergence analysis of \texttt{DeFL} on \texttt{Multi-Krum} is similar to that on \texttt{Krum}, therefore, we omit it in Section~\ref{Section:Analysis:BFT} and Section~\ref{Section:Analysis:Convergence} due to the page limit. Note that we have not provided theoretical analysis when the data are non-i.i.d., however, we empirically evaluate this scenario on public datasets in Section~\ref{Section:Evaluation}.

\subsection{Overhead}
\textbf{Network bandwidth.} The communication complexity per view is $O(n)$ in HotStuff~\cite{DBLP:conf/podc/YinMRGA19}. In a training round, an honest node would commit an \texttt{UPD} transaction and an \texttt{AGG} transaction. As the size of weights $M$ is much larger than that of $\mathit{id}$ or $\mathit{round\_id}$, we only consider $M$ here. Therefore, it takes $O(Mn)$ network bandwidth to synchronize weights of the current round on an honest node. As there are $O(n)$ honest nodes, the overall network bandwidth of $T$ training rounds is $O(MTn^2)$.

\textbf{Storage.} Thanks to the decoupling-storage-and-consensus design, \texttt{DeFL} maintains and caches weights of only some constant $\tau\geq 2$ training rounds from $n$ nodes, instead of maintaining the consistency of all history weights. Therefore, the overall storage complexity is $M\tau n$, regardless of the number of training rounds.

\summary{When $n\ge 3f+3$, and the data are i.i.d., \texttt{DeFL} is BFT and reaches convergence in a bounded time period. The network bandwidth of $T$ training rounds is $O(MTn^2)$ and the storage complexity is $M\tau n$, where $M$ is the size of weights and \texttt{DeFL} caches $\tau$ training rounds in storage.}

%% file: Sections/Evaluation.tex
In this section, we introduce the experimental setup and measure the fault tolerance and scalability of \texttt{DeFL}.

\subsection{Experiement Setup} \label{Section:Evaluation:Setup}
\textbf{Datasets.} We use CIFAR-10~\footnote{https://www.cs.toronto.edu/$\sim$kriz/cifar.html} for image classification under MIT License. It consists of 60,000 32x32 color images in 10 classes, with 6,000 images per class. There are 50,000 training images and 10,000 test images. We use Sentiment140~\footnote{http://help.sentiment140.com/for-students} for sentiment analysis. The license of Sentiment140 allows academic use without commercial purposes. Because the size of its official testing set (498) is much smaller than its official training set (1,600,000), the test accuracy would be unstable, especially when this dataset is separated into many nodes. Therefore, we manually remove the labels of 160,000 (10\%) samples in the official training set and treat them as the new testing set. Therefore, there are 1,440,000 training sentences and 160,000 testing sentences. The number of positive reviews and negative ones are identical. For CIFAR-10 and Sentiment140, we follow related work~\cite{DBLP:journals/corr/abs-1909-06335, DBLP:conf/nips/LinKSJ20, DBLP:conf/icml/ZhuHZ21} to model non-i.i.d. data using a Dirichlet distribution $\texttt{Dir}(\alpha)$, in which a smaller $\alpha$ indicates higher data heterogeneity
. We choose $\alpha=1$ to create CIFAR-noniid and Sentiment-noniid datasets. Both datasets do not contain personally identifiable information or offensive content. Note that the experimental results of Sentiment140 and Sentiment-noniid are similar to those of CIFAR-10 and CIFAR-noniid and are present in Section~\ref{Section:Appendix} due to the page limit.

\textbf{Models.} For CIFAR-10 and CIFAR-noniid, we use Dense Convolutional Network (DenseNet)
. The depth of DenseNet is 100 and the growth rate is 12. The batch size is 32. The learning rate is 1e-3. For Sentiment140 and Sentiment-noniid, we use attention-based Bidirectional Long Short-Term Memory (Bi-LSTM)
. The length, iteration number, and window size of Word2Vec embeddings are 300, 32, and 7, respectively.  
The number of Bi-LSTM units is 128. The batch size is 1,024. The dropout rate of the fully connected layer and the attention layer is 0.15.

\textbf{Environments.} We deploy \texttt{DeFL} and baselines on 4-10 instances, each equipped with an NVIDIA Tesla K80 GPU with 12GB memory, 3 Xeon E5-2678 v3 CPUs, and 8GB RAM. 
Each setting is repeated 10 times to take the average as reported results.

\textbf{Baselines.} We carefully choose 3 aforementioned open-source representative baselines to compare with \texttt{DeFL}, i.e. FL~\cite{DBLP:conf/aistats/McMahanMRHA17}, Swarm Learning (SL)~\cite{warnat2021swarm}, and \texttt{Biscotti}~\cite{DBLP:journals/tpds/ShayanFYB21}. FL has no defense against poisoning attacks. SL uses a blockchain to elect a leader. \texttt{Biscotti} defends against poisoning attacks via \texttt{Multi-Krum}~\cite{DBLP:conf/nips/BlanchardMGS17} and uses a blockchain to store and maintain the consistency of weights.

\subsection{Fault Tolerance}\label{Section:RQ1}
\begin{table}[htbp]
  \caption{Accuracy on different threat models}
  \label{RQ1.1}
  \begin{tabular}{m{0.24\columnwidth}m{0.06\columnwidth}m{0.06\columnwidth}m{0.06\columnwidth}m{0.06\columnwidth}m{0.06\columnwidth}m{0.06\columnwidth}m{0.06\columnwidth}m{0.06\columnwidth}}
    \toprule
    & \multicolumn{4}{c}{CIFAR-10} & \multicolumn{4}{c}{CIFAR-noniid} \\
    \tc{Attack}&\tc{FL}&\tc{SL}&\tc{Biscotti}&\tc{DeFL}&\tc{FL}&\tc{SL}&\tc{Biscotti}&\tc{DeFL}\\
    \midrule
    No & \tr{0.924} & \tr{\textbf{0.926}} & \tr{0.891} & \tr{0.899} & \tr{0.922} & \tr{\textbf{0.925}} & \tr{0.840} & \tr{0.836}\\
    Gaussian ($\sigma$=0.03) & \tr{\textbf{0.905}} & \tr{0.904} & \tr{0.887} & \tr{0.888} & \tr{0.922} & \tr{\textbf{0.924}} & \tr{0.891} & \tr{0.893}\\
    Gaussian ($\sigma$=1.00) & \tr{0.184} & \tr{0.197} & \tr{\textbf{0.899}} & \tr{0.894} & \tr{0.345} & \tr{0.338} & \tr{0.872} & \tr{\textbf{0.876}}\\
    Sign-flipping ($\sigma$=-1.0) & \tr{0.837} & \tr{0.843} & \tr{0.880} & \tr{\textbf{0.885}} & \tr{0.799} & \tr{0.803} & \tr{\textbf{0.888}} & \tr{0.883}\\
    Sign-flipping ($\sigma$=-2.0) & \tr{0.453} & \tr{0.456} & \tr{0.890} & \tr{\textbf{0.893}} & \tr{0.423} & \tr{0.421} & \tr{0.878} & \tr{\textbf{0.881}}\\
    Sign-flipping ($\sigma$=-4.0) & \tr{0.126} & \tr{0.136} & \tr{\textbf{0.896}} & \tr{0.893} & \tr{0.164} & \tr{0.175} & \tr{0.866} & \tr{\textbf{0.873}}\\
    Label-flipping & \tr{\textbf{0.894}} & \tr{0.893} & \tr{0.889} & \tr{0.890} & \tr{\textbf{0.890}} & \tr{0.884} & \tr{0.872} & \tr{0.876}\\
  \bottomrule
\end{tabular}
\end{table}
As mentioned in Section~\ref{Section:Methodology:Threat_model}, we measure 3 types of poisoning attacks with different attack factors on 1 of 4 nodes. As shown in Table~\ref{RQ1.1}, \texttt{FedAvg}-based approaches (FL, SL) share similar accuracy while \texttt{Multi-Krum}-based approaches (\texttt{Biscotti}, \texttt{DeFL}) share similar accuracy. When there is no or a mild attack, i.e. Gaussian attack ($\sigma$=0.03) and label-flipping attack, the accuracy of \texttt{FedAvg}-based approaches is slightly higher than that of \texttt{Multi-Krum}-based ones. The reason is that \texttt{Multi-Krum} filters outlier weights that might be trained by honest nodes and are supposed to be aggregated. This is also why the accuracy of \texttt{Multi-Krum}-based approaches on CIFAR-noniid is lower than that of CIFAR-10. However, when there is a relatively severe attack, the accuracy of \texttt{Multi-Krum}-based approaches is significantly higher than that of \texttt{FedAvg}-based ones. This indicates that \texttt{Multi-Krum} effectively detects poisoned weights and aggregates with \textit{correct} weights from honest nodes.

\begin{table}[htbp]
  \caption{Accuracy on CIFAR-noniid with sign-flipping attack ($\sigma$=-2.0)}
  \label{RQ1.2}
  \centering
  \begin{tabular}{m{0.15\columnwidth}m{0.1\columnwidth}m{0.1\columnwidth}m{0.1\columnwidth}m{0.1\columnwidth}}
    \toprule
    \tc{Attack}&\tc{FL}&\tc{SL}&\tc{Biscotti}&\tc{DeFL}\\
    \midrule
    4+0 ($\beta$=0.00) & \tr{0.922} & \tr{\textbf{0.925}} & \tr{0.840} & \tr{0.836}\\
    3+1 ($\beta$=0.25) & \tr{0.423} & \tr{0.421} & \tr{0.878} & \tr{\textbf{0.881}}\\
    7+0 ($\beta$=0.00) & \tr{\textbf{0.891}} & \tr{0.890} & \tr{0.823} & \tr{0.825}\\
    6+1 ($\beta$=0.14) & \tr{0.717} & \tr{0.722} & \tr{\textbf{0.851}} & \tr{0.850}\\
    5+2 ($\beta$=0.29) & \tr{0.380} & \tr{0.369} & \tr{0.865} & \tr{\textbf{0.874}}\\
    10+0 ($\beta$=0.00) & \tr{\textbf{0.883}} & \tr{0.881} & \tr{0.832} & \tr{0.826}\\
    9+1 ($\beta$=0.10) & \tr{0.775} & \tr{0.779} & \tr{\textbf{0.845}} & \tr{0.842}\\
    8+2 ($\beta$=0.20) & \tr{0.631} & \tr{0.634} & \tr{0.850} & \tr{\textbf{0.855}}\\
    7+3 ($\beta$=0.30) & \tr{0.358} & \tr{0.353} & \tr{0.874} & \tr{\textbf{0.878}}\\
  \bottomrule
\end{tabular}
\end{table}
We choose sign-flipping attack ($\sigma$=-2.0) on CIFAR-noniid to measure the accuracy under different Byzantine rates $\beta$, as shown in Table~\ref{RQ1.2}. We scale \texttt{DeFL} on 4, 7, 10 nodes, and "$a$+$b$" means there are $a$ honest nodes and $b$ Byzantine node(s). 
As $\beta$ raises, the accuracy of \texttt{FedAvg}-based approaches drops dramatically, while the accuracy of \texttt{Multi-Krum}-based ones remains stable and significantly higher.

\summary{Under most attacks, the accuracy of \texttt{DeFL} and \texttt{Biscotti} is significantly higher than that of FL and SL, no matter the Byzantine rate. However, when there is no or a mild attack, the accuracy of \texttt{DeFL} and \texttt{Biscotti} is slightly lower than that of FL or SL.}

\subsection{Scalability}\label{Section:RQ2}
\begin{figure}[htbp]
\centering
  \begin{tabular}[t]{lll}
    \subfigure{
            \includegraphics[width=0.33\textwidth,valign=t]{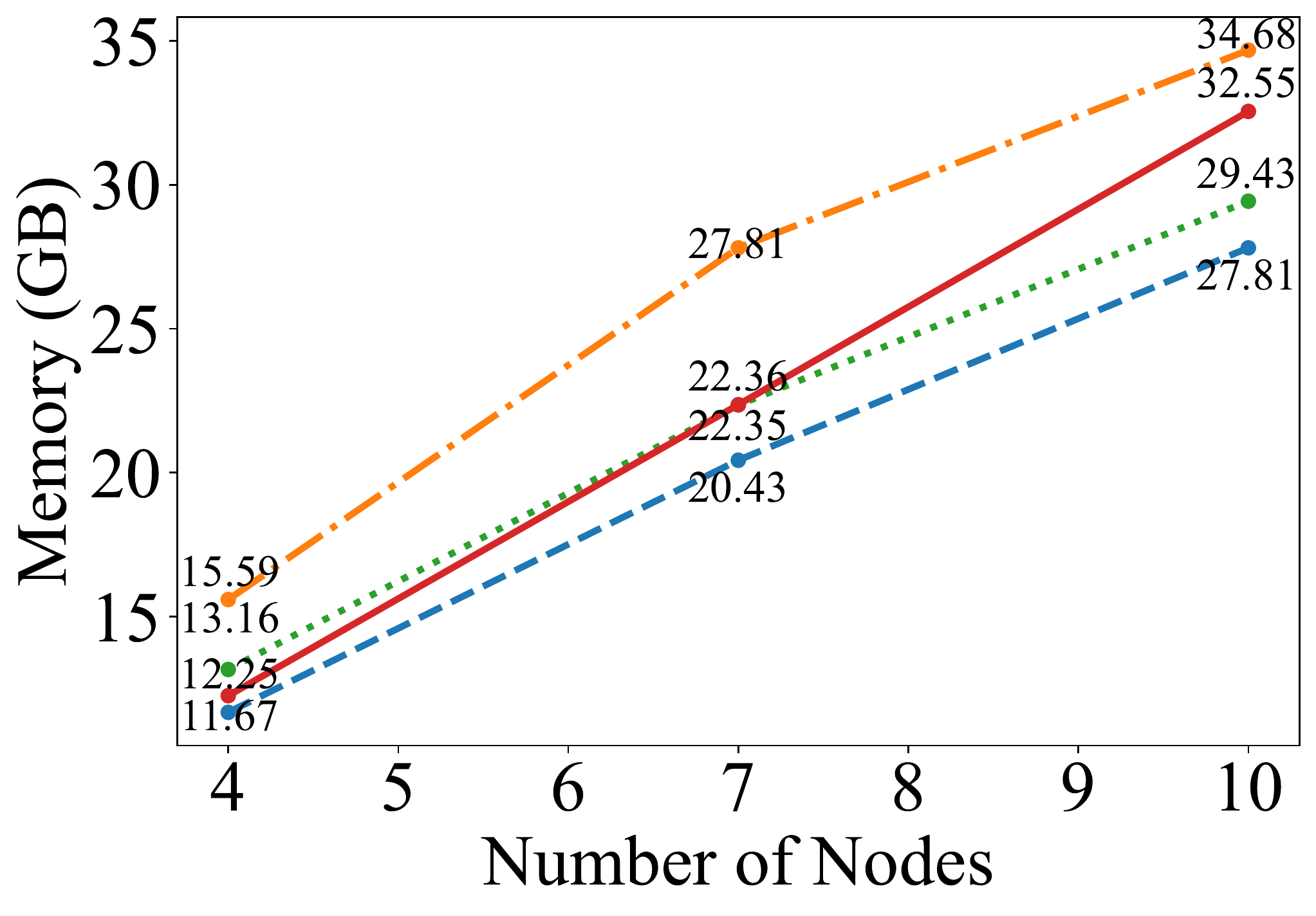}
    }
    &
    \subfigure{
            \includegraphics[width=0.33\textwidth,valign=t]{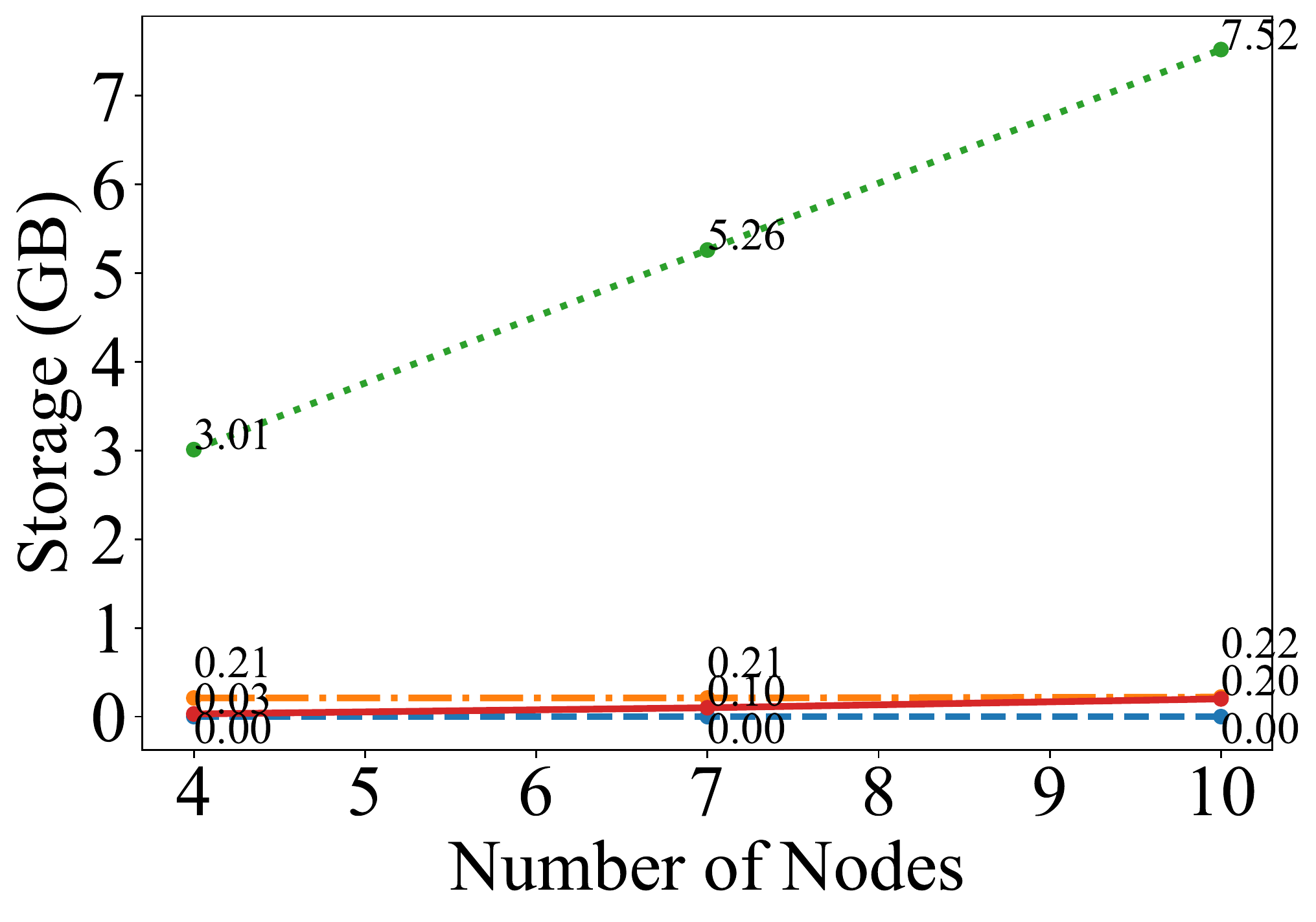}
    }
    &
    \subfigure{
            \includegraphics[width=0.12\textwidth,valign=t]{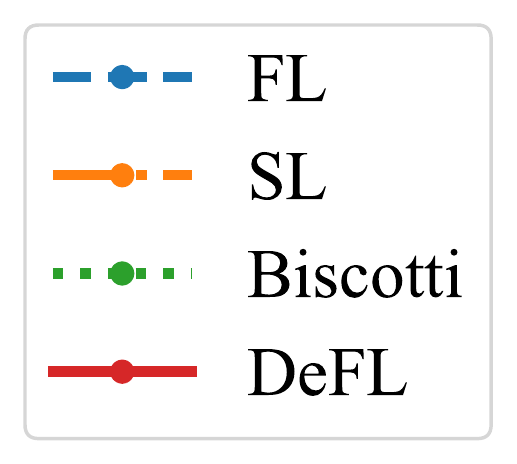}
    }
    \\
    \subfigure{
            \includegraphics[width=0.33\textwidth,valign=t]{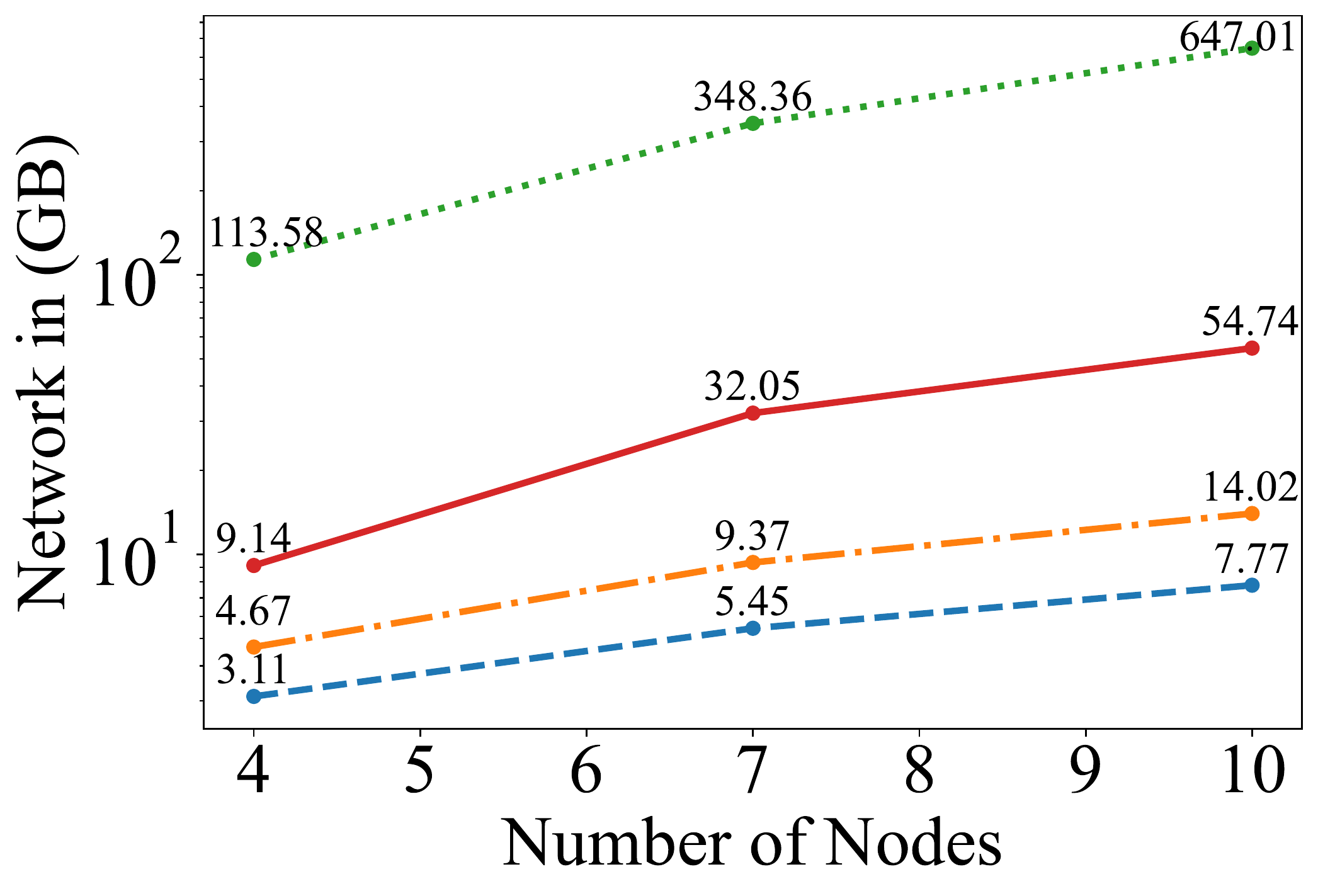}
    }
    &
    \subfigure{
            \includegraphics[width=0.33\textwidth,valign=t]{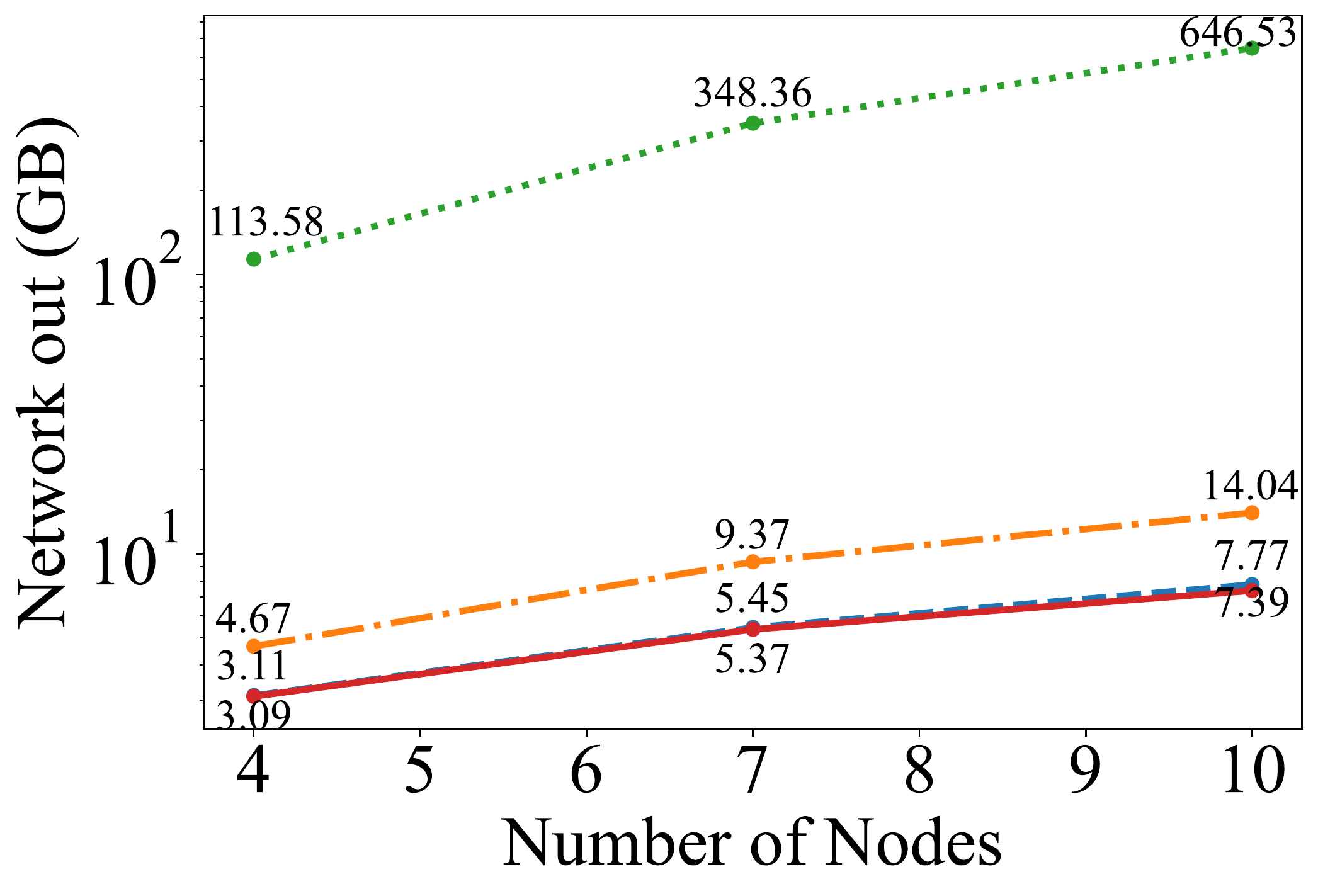}
    }
    &
    \\
\end{tabular}
\caption{Overhead of different scales on CIFAR-noniid}
\label{RQ2}
\end{figure}

We scale \texttt{DeFL} and baselines to 4, 7, 10 nodes and measure computation, storage, and network overhead. As shown in Figure~\ref{RQ2}, as the number of nodes increases, the RAM usage of \texttt{DeFL} and other baselines increases linearly. The RAM usage of \texttt{DeFL} is close to that of \texttt{Biscotti} and FL, and lower than that of SL. The GPU memory usage of \texttt{DeFL} and other baselines is always identical
. In terms of storage overhead, we measure the storage usage of only the blockchain for fairness. The storage of \texttt{DeFL} is close to that of FL and SL (nearly 0 GB), and significantly lower than that of \texttt{Biscotti}, thanks to the decoupling-storage-and-consensus design. In terms of network overhead, we change the y-axis into a logarithmic axis for clear presentation. The receiving bandwidth of SL and FL is linear to the number of nodes, while that of \texttt{DeFL} and \texttt{Biscotti} is quadratic to the number of nodes. Besides, the receiving bandwidth of \texttt{DeFL} is much lower than that of \texttt{Biscotti}, though higher than that of FL and SL. Thanks to the shared memory pool, the sending bandwidth of \texttt{DeFL} is linear to the number of nodes and similar to (even slightly lower than) that of FL, while that of other baselines is similar to their receiving bandwidth.

\summary{The computational overhead of \texttt{DeFL} is similar to that of baselines. The storage overhead of \texttt{DeFL} is almost 0 GB. Although the network overhead of \texttt{DeFL} increases quadratically, it is up to 12x lower than that of \texttt{Biscotti}, which achieves similar accuracy to \texttt{DeFL} under most attacks.}

%% file: Sections/Discussion.tex
In this section, we discuss some threats to our approach, including limited scalability and competitive literature.

\subsection{Extend to Cross-device Federated Learning}
Note that we limit the scope to cross-silo FL, which makes the following assumptions and results acceptable, (i) partially synchronous communication, (ii) independently local training, (iii) $O(MTn^2)$ network bandwidth. (i) is the assumption of \texttt{HotStuff}~\cite{DBLP:conf/podc/YinMRGA19}, however, in cross-device FL, where device heterogeneity is relatively high, we will consider asynchronous federated learning (AFL)~\cite{DBLP:conf/pkdd/SpragueJSCNDK18} to weaken this assumption in future work. AFL allows the central server to aggregate weights as soon as it receives a local model. (ii) is the assumption of most FL approaches, however, in cross-device FL, where devices are resource-constrained, we will consider on-device training optimizations, like pruning~\cite{DBLP:conf/nips/HanPTD15}, quantization~\cite{DBLP:journals/corr/HanMD15}, and knowledge distillation~\cite{DBLP:journals/corr/HintonVD15}. (iii) is the consequence of decentralized aggregation on each node. It could optimize the network bandwidth to reduce the number of synchronizations by designing consensus specialized for FL. For example, \texttt{DeFL} enables updates in one training round to be recorded in a particular sequence, however, this feature is not necessary for FL. We will consider asynchronous Byzantine Atomic Broadcast (BAB) protocols, such as \texttt{DAG-Rider}~\cite{DBLP:conf/podc/KeidarKNS21}, to slack the sequence constraint.

\subsection{Difference from Byzantine Fault Tolerant Federated Learning}
There are many efforts in defending Byzantine attacks in FL and they can be categorized into proactive and reactive defenses~\cite{DBLP:journals/fgcs/MothukuriPPHDS21}. Proactive defenses include knowledge distillation~\cite{DBLP:journals/corr/abs-1910-03581}, pruning~\cite{DBLP:journals/corr/abs-1909-12326}, moving target defense~\cite{DBLP:conf/isi/ColbaughG13}, data sanitization~\cite{DBLP:conf/sp/CretuSLSK08}, and federated multi-task learning~\cite{DBLP:conf/nips/SmithCST17}. Reactive defenses include Sniper~\cite{DBLP:conf/icpads/CaoCLLS19}, anomaly detection~\cite{DBLP:conf/nips/BlanchardMGS17}, Foolsgold~\cite{DBLP:journals/corr/abs-1808-04866}. However, conventional BFT FL approaches are based on the assumption that there exists an honest central parameter server. When the central server is eliminated in \texttt{DeFL}, most of them become invalid.

%% file: Sections/Conclusion.tex
In this paper, we propose a novel decentralized weight aggregation framework for federated learning  (\texttt{DeFL}), where weights are aggregated on each node and weights of only the current training round are maintained and synchronized. To tackle the weight poisoning and round number inconsistency challenges, \texttt{DeFL} enables aggregating \textit{correct} weights from honest nodes based on \texttt{Multi-Krum} and ensures the consistency of the round number and weights based on \texttt{HotStuff}. We theoretically analyze the Byzantine fault tolerance, convergence, and overhead of \texttt{DeFL} and evaluate these results over two public datasets. The experimental results validate the effectiveness of \texttt{DeFL} in defending against common threat models with minimal accuracy loss, as well as show the efficiency of \texttt{DeFL} in storage and network usage. In future work, we will leverage asynchronous FL~\cite{DBLP:conf/pkdd/SpragueJSCNDK18} and asynchronous BAB protocols~\cite{DBLP:conf/podc/KeidarKNS21} to extend \texttt{DeFL} to cross-device FL.

%% file: Sections/Appendix.tex
\section{Extended Results for Sentiment140} \label{Section:Appendix}
In this section, we present extended results for Sentiment140 and Sentiment-noniid datasets in fault tolerance and scalability.

\subsection{Fault Tolerance}\label{Section:Appendix:RQ1}
\begin{table}[htbp]
  \caption{Accuracy on different threat models}
  \label{Appendix:RQ1.1}
  \begin{tabular}{m{0.24\columnwidth}m{0.06\columnwidth}m{0.06\columnwidth}m{0.06\columnwidth}m{0.06\columnwidth}m{0.06\columnwidth}m{0.06\columnwidth}m{0.06\columnwidth}m{0.06\columnwidth}}
    \toprule
    & \multicolumn{4}{c}{Sentiment140} & \multicolumn{4}{c}{Sentiment-noniid} \\
    \tc{Attack}&\tc{FL}&\tc{SL}&\tc{Biscotti}&\tc{DeFL}&\tc{FL}&\tc{SL}&\tc{Biscotti}&\tc{DeFL}\\
    \midrule
    No & \tr{0.745} & \tr{\textbf{0.746}} & \tr{0.744} & \tr{\textbf{0.746}} & \tr{0.700} & \tr{0.699} & \tr{\textbf{0.701}} & \tr{0.698}\\
    Gaussian ($\sigma$=0.03) & \tr{0.745} & \tr{0.743} & \tr{\textbf{0.746}} & \tr{\textbf{0.746}} & \tr{0.699} & \tr{\textbf{0.701}} & \tr{0.700} & \tr{0.699}\\
    Gaussian ($\sigma$=1.00) & \tr{0.737} & \tr{0.736} & \tr{0.745} & \tr{\textbf{0.747}} & \tr{0.537} & \tr{0.534} & \tr{\textbf{0.701}} & \tr{0.699}\\
    Sign-flipping ($\sigma$=-1.0) & \tr{0.736} & \tr{0.738} & \tr{\textbf{0.749}} & \tr{0.747} & \tr{0.685} & \tr{0.686} & \tr{0.698} & \tr{\textbf{0.699}}\\
    Sign-flipping ($\sigma$=-2.0) & \tr{0.725} & \tr{0.722} & \tr{\textbf{0.750}} & \tr{0.748} & \tr{0.699} & \tr{\textbf{0.700}} & \tr{0.699} & \tr{\textbf{0.700}}\\
    Sign-flipping ($\sigma$=-4.0) & \tr{0.655} & \tr{0.659} & \tr{0.745} & \tr{\textbf{0.748}} & \tr{0.508} & \tr{0.510} & \tr{0.697} & \tr{\textbf{0.700}}\\
    Label-flipping & \tr{0.719} & \tr{0.720} & \tr{\textbf{0.746}} & \tr{\textbf{0.746}} & \tr{0.698} & \tr{0.699} & \tr{\textbf{0.701}} & \tr{0.700}\\
  \bottomrule
\end{tabular}
\end{table}
As mentioned in Section~\ref{Section:Methodology:Threat_model}, we measure 3 types of poisoning attacks with different attack factors on 1 of 4 nodes. As shown in Table~\ref{Appendix:RQ1.1}, \texttt{FedAvg}-based approaches (FL, SL) share similar accuracy while \texttt{Multi-Krum}-based approaches (\texttt{Biscotti}, \texttt{DeFL}) share similar accuracy. In most settings, especially when the attack is severe, the accuracy of \texttt{Multi-Krum}-based approaches is higher than that of \texttt{FedAvg}-based ones. This indicates that \texttt{Multi-Krum} effectively detects poisoned weights and aggregates with \textit{correct} weights from honest nodes.

\begin{table}[htbp]
  \caption{Accuracy on Sentiment-noniid with Gaussian attack ($\sigma$=1.00)}
  \label{Appendix:RQ1.2}
  \centering
  \begin{tabular}{m{0.15\columnwidth}m{0.1\columnwidth}m{0.1\columnwidth}m{0.1\columnwidth}m{0.1\columnwidth}}
    \toprule
    \tc{Attack}&\tc{FL}&\tc{SL}&\tc{Biscotti}&\tc{DeFL}\\
    \midrule
    4+0 ($\beta$=0.00) & \tr{0.700} & \tr{0.699} & \tr{\textbf{0.701}} & \tr{0.698}\\
    3+1 ($\beta$=0.25) & \tr{0.537} & \tr{0.539} & \tr{\textbf{0.700}} & \tr{0.699}\\
    7+0 ($\beta$=0.00) & \tr{\textbf{0.701}} & \tr{0.700} & \tr{0.700} & \tr{\textbf{0.701}}\\
    6+1 ($\beta$=0.14) & \tr{0.624} & \tr{0.622} & \tr{\textbf{0.701}} & \tr{0.700}\\
    5+2 ($\beta$=0.29) & \tr{0.573} & \tr{0.570} & \tr{0.700} & \tr{\textbf{0.701}}\\
    10+0 ($\beta$=0.00) & \tr{\textbf{0.701}} & \tr{0.699} & \tr{0.700} & \tr{\textbf{0.701}}\\
    9+1 ($\beta$=0.10) & \tr{0.656} & \tr{0.660} & \tr{\textbf{0.702}} & \tr{0.701}\\
    8+2 ($\beta$=0.20) & \tr{0.633} & \tr{0.631} & \tr{0.701} & \tr{\textbf{0.702}}\\
    7+3 ($\beta$=0.30) & \tr{0.601} & \tr{0.604} & \tr{0.700} & \tr{\textbf{0.702}}\\
  \bottomrule
\end{tabular}
\end{table}
We choose Gaussian attack ($\sigma$=1.00) on Sentiment-noniid to measure the accuracy under different Byzantine rates $\beta$, as shown in Table~\ref{Appendix:RQ1.2}. We scale \texttt{DeFL} on 4, 7, 10 nodes, and "$a$+$b$" means there are $a$ honest nodes and $b$ Byzantine node(s). 
When the number of nodes is fixed, as $\beta$ raises, the accuracy of \texttt{FedAvg}-based approaches drops dramatically, while the accuracy of \texttt{Multi-Krum}-based ones remains stable and significantly higher. Besides, when $\beta$ keeps stable or even slightly raises, as the number of nodes increases, the accuracy of \texttt{FedAvg}-based approaches raises. The reason could be that the higher ensemble property of more nodes contributes to better generalizability and robustness, however, it is beyond the scope of this paper.

\summary{Under most attacks, the accuracy of \texttt{DeFL} and \texttt{Biscotti} is significantly higher than that of FL and SL, no matter the Byzantine rate.}

\subsection{Scalability}\label{Section:Appendix:RQ2}
\begin{figure}[htbp]
\centering
  \begin{tabular}[t]{lll}
    \subfigure{
            \includegraphics[width=0.33\textwidth,valign=t]{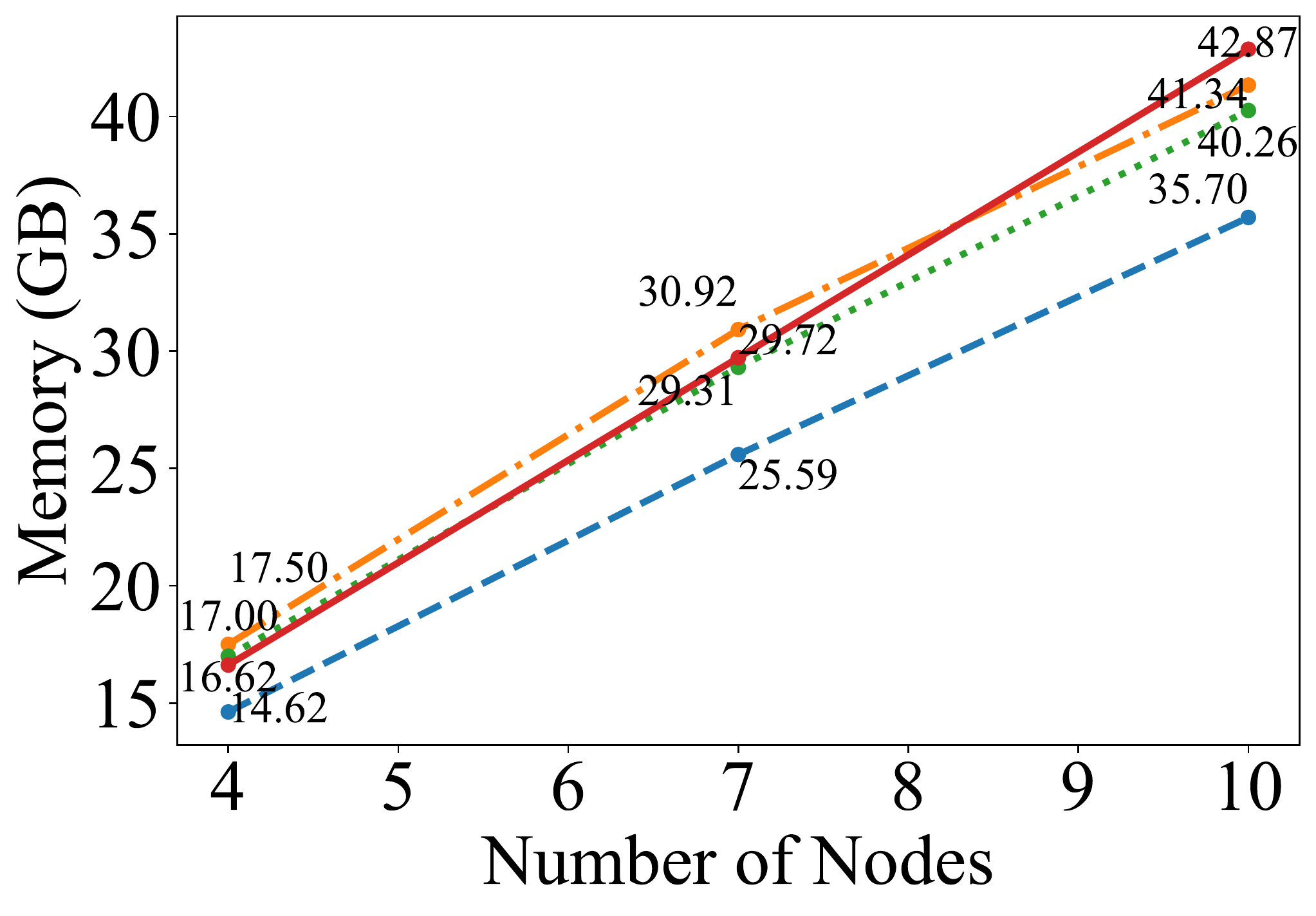}
    }&
    \subfigure{
            \includegraphics[width=0.33\textwidth,valign=t]{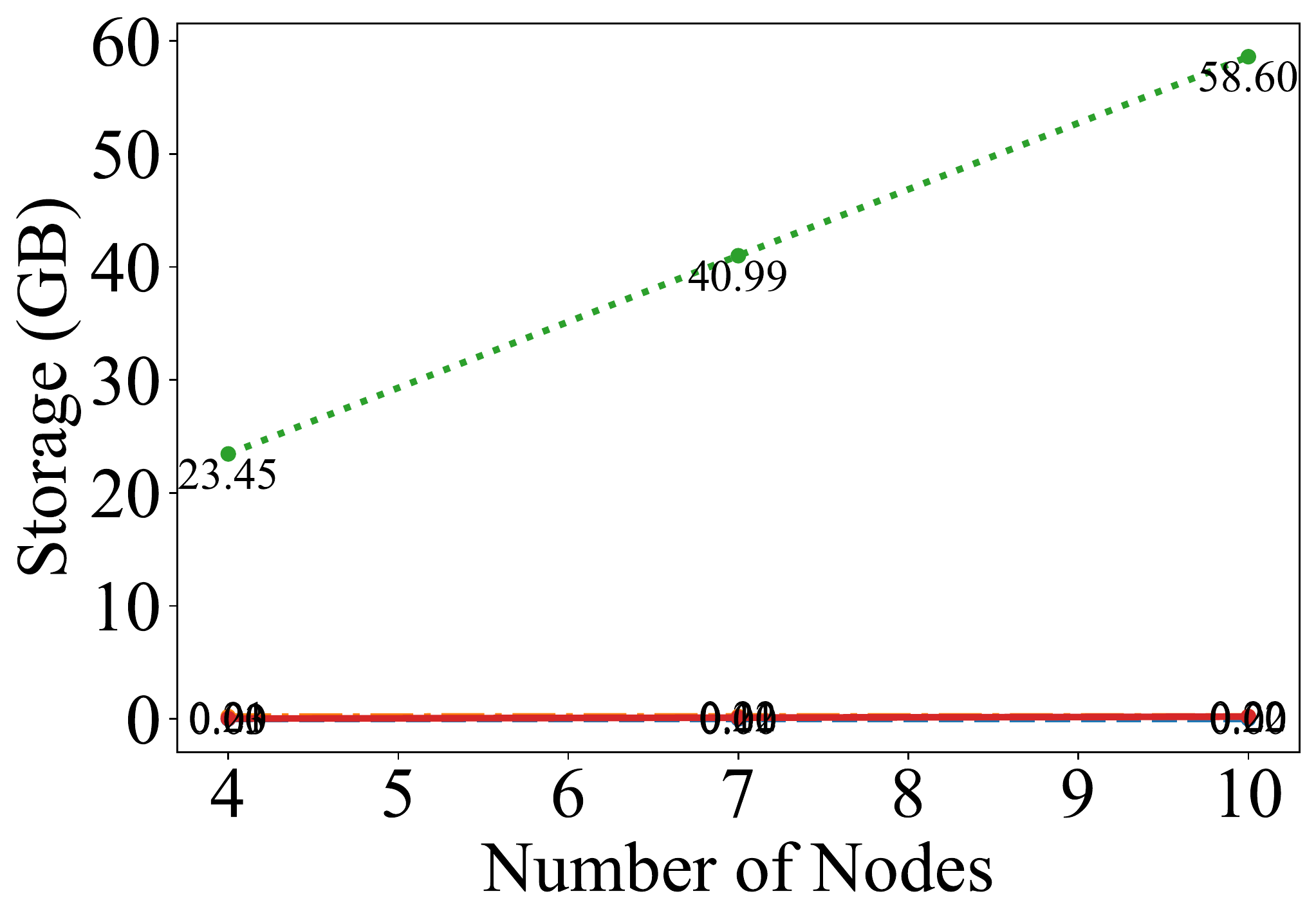}
    }
    &
    \subfigure{
            \includegraphics[width=0.12\textwidth,valign=t]{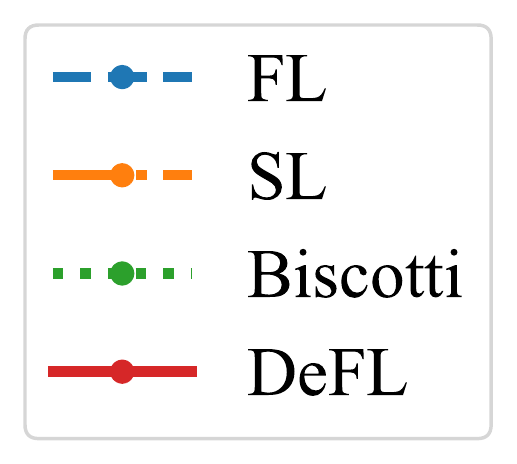}
    }
    \\
    \subfigure{
            \includegraphics[width=0.33\textwidth,valign=t]{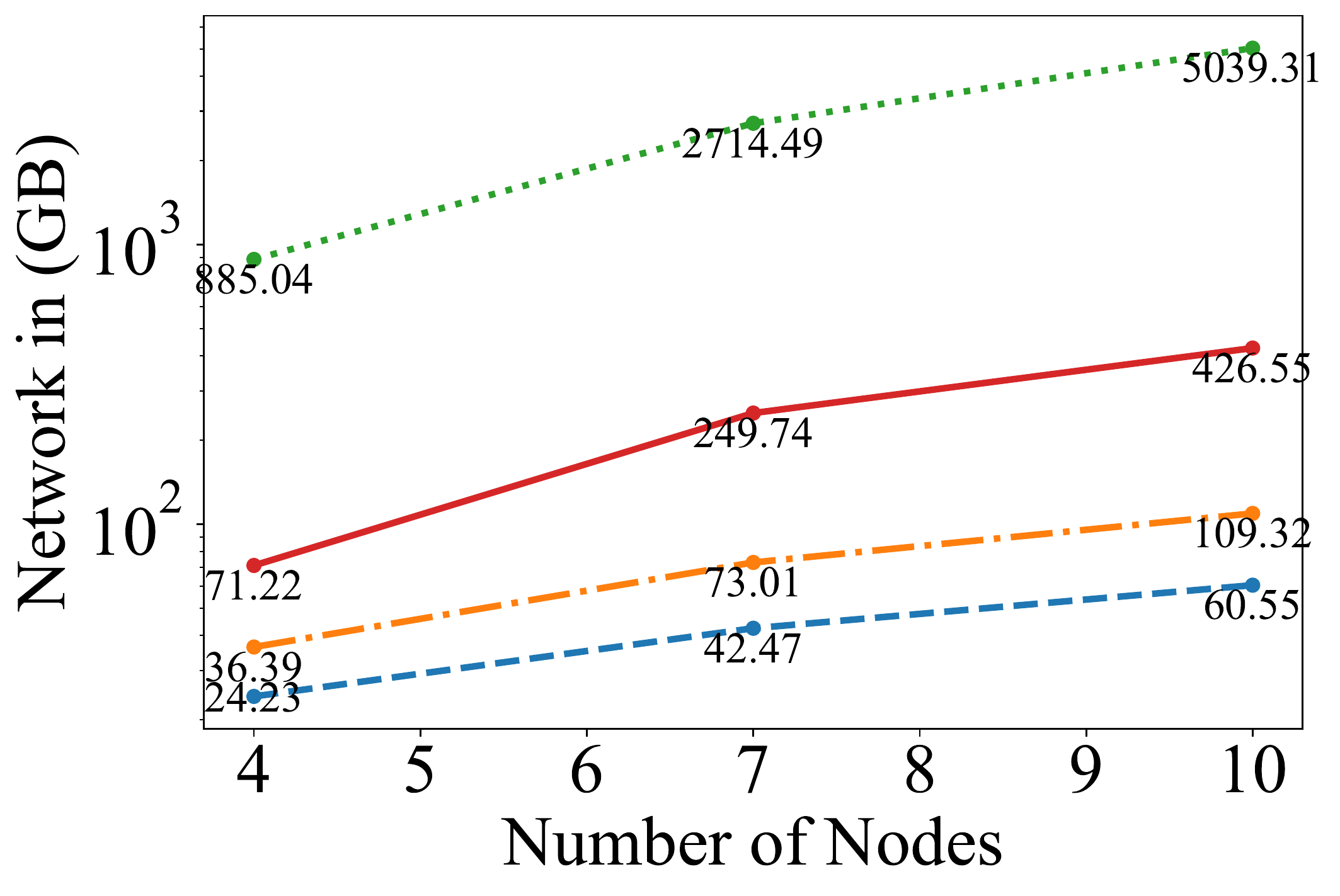}
    }
    &
    \subfigure{
            \includegraphics[width=0.33\textwidth,valign=t]{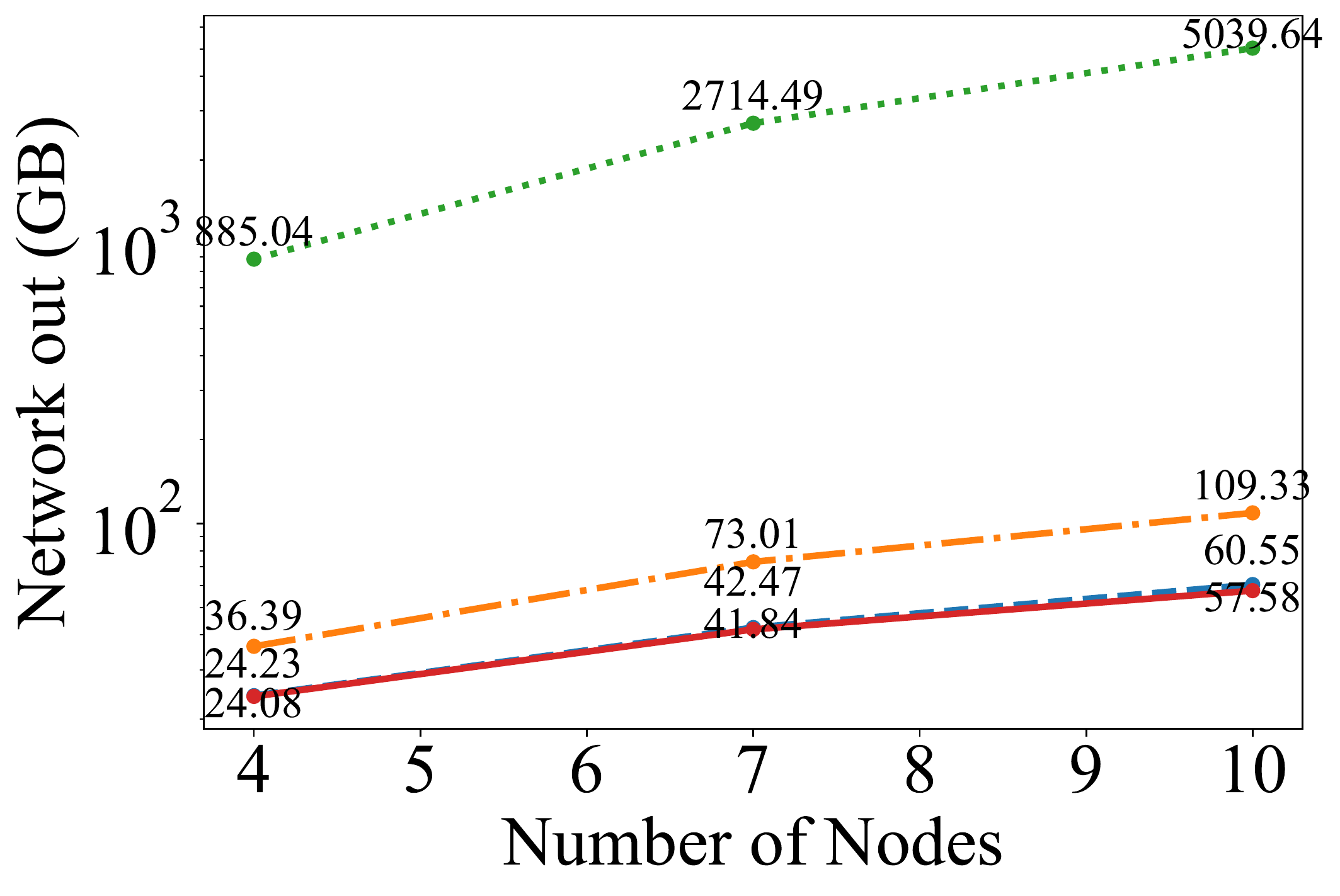}
    }
    &
    \\
\end{tabular}
\caption{Overhead of different scales on Sentiment-noniid}
\label{Appendix:RQ2}
\end{figure}

We scale \texttt{DeFL} and baselines to 4, 7, 10 nodes and measure computation, storage, and network overhead. As shown in Figure~\ref{Appendix:RQ2}, as the number of nodes increases, the growth trends of overhead of \texttt{DeFL} and baselines are similar to those on CIFAR-noniid, as mentioned in Section~\ref{Section:RQ2}.

\summary{The computational overhead of \texttt{DeFL} is similar to that of baselines. The storage overhead of \texttt{DeFL} is almost 0 GB. Although the network overhead of \texttt{DeFL} increases quadratically, it is up to 12x lower than that of \texttt{Biscotti}, which achieves similar accuracy to \texttt{DeFL} under most attacks.}